\documentclass[sigconf]{acmart}

\renewcommand\footnotetextcopyrightpermission[1]{} 

\usepackage{algorithm}
\usepackage{algorithmic}
\usepackage{cleveref}
\usepackage{natbib}
\usepackage{float}
\AtBeginDocument{%
  \providecommand\BibTeX{{%
    \normalfont B\kern-0.5em{\scshape i\kern-0.25em b}\kern-0.8em\TeX}}}

\begin{document}

\title{Impact of Accuracy on Model Interpretations}

\author{Brian Liu}
\affiliation{%
  \institution{Cornell University}
}
\email{bl462@cornell.edu}
\author{Madeleine Udell}
\affiliation{%
  \institution{Cornell University}
}
\email{udell@cornell.edu}


\begin{abstract}
Model interpretations are often used in practice to extract real world insights from machine learning models. These interpretations have a wide range of applications; they can be presented as business recommendations or used to evaluate model bias. It is vital for a data scientist to choose trustworthy interpretations to drive real world impact. Doing so requires an understanding of how the accuracy of a model impacts the quality of standard interpretation tools.

In this paper, we will explore how a model's predictive accuracy affects interpretation quality. We propose two metrics to quantify the quality of an interpretation and design an experiment to test how these metrics vary with model accuracy. We find that for datasets that can be modeled accurately by a variety of methods, simpler methods yield higher quality interpretations. We also identify which interpretation method works the best for lower levels of model accuracy.

\end{abstract}



\maketitle
\pagestyle{plain}
\section{Introduction}

In applied data science, high performing black box models are increasingly being used to solve complicated real-world problems. While these models can be incredibly accurate, they lack interpretability. It is very difficult to look at these models and deduce the relationship between their features and responses. Interpretable machine learning aims to address this issue by developing interpretation algorithms that give insight on how complex models work. In recent years, interpretation algorithms such as LIME \cite{DBLP:journals/corr/RibeiroSG16} and SHAP \cite{DBLP:journals/corr/LundbergL17} have become commonplace in a data scientist's toolbox. As topics such as fairness and transparency in AI/ML become more important, the necessity of interpretability in machine learning will continue to grow.

 Often times, data scientists are tasked to determine causal relationships from models with less than ideal accuracy. How does a model's accuracy impact the trustworthiness of its interpretations? Understanding how predictive accuracy impacts interpretation quality would make such analyses more trustworthy.

This paper examines how model interpretations, specifically feature importance rankings, change when the accuracy of a binary classification model decreases. We test, using computational experiments, how various interpretable models and interpretation algorithms fare when a model's accuracy is artificially reduced. From our results, we can determine which interpretation methods perform best for a given level of model accuracy.

\subsection{Main Contributions}
The main contributions of this paper follow.
\begin{itemize}
    \item We propose two new metrics to quantify the quality of model interpretations.
    \item We design an experiment that tests how interpretation quality varies with model accuracy.
    \item We present a list of recommendations to data scientists on which interpretation methods should be used for different levels of model accuracy.
\end{itemize}

\subsection{Paper layout}
Our paper will first discuss the related literature concerning machine learning interpretability and interpretation quality. We then present a list of interpretation methods we consider, as well as our methodology for testing these methods. We explain the two metrics we create to quantify interpretation quality and then discuss our experiment's results. We conclude with our list of recommendations to data scientists on which interpretation methods to use.

\section{Background}
In the following, we will discuss literature relevant to our work along with the established model interpretation methods we use.

\subsection{Related Work}
In this section, we demonstrate how our work builds upon existing literature. 

Murdoch and Singh define the PDR (Predictive, Descriptive, Relevant) framework for evaluating interpretation quality \cite{1901.04592}. The framework states that an interpretable model should strive to have high predictive accuracy (the test accuracy of the model) and high descriptive accuracy (the degree the interpretation method captures relationships learned by the model). We will build upon Murdoch and Singh's work by examining how predictive accuracy impacts descriptive accuracy in an individual model. We aim to the quantify the effect predictive accuracy has on descriptive accuracy.

We will evaluate the descriptive accuracy of an interpretation through using functional proxies, a concept proposed by Doshi-Velez and Kim \cite{1702.08608}. Functional proxies use a formalized definition of descriptive accuracy as a metric to evaluate the quality of an interpretation. In the context of our work, we will decompose descriptive accuracy into the combination of interpretation trueness and interpretation stability/precision. This follows the definition of accuracy set by ISO 5725 \cite{ISO5725}. Both components are necessary for a interpretation to accurately capture the relationship learned by a model, as shown in figure 1. 

We define interpretation trueness as the probability of obtaining the ground-truth feature ranks when interpreting a model. Since the ground-truth feature rank is unknown, we will define an upper bound on this probability. Our definition of interpretation stability builds upon Yu's definition that statistical stability holds when conclusions are robust to perturbations in the data \cite{1310.0150}. Since we are interested in feature importance rankings, a stable interpretation method should give similar feature rankings when fit on bootstrapped data. Our formalized approach to bound interpretation trueness and measure  stability is detailed in our methodology section.

 \begin{figure}[h]
\includegraphics[width=8 cm]{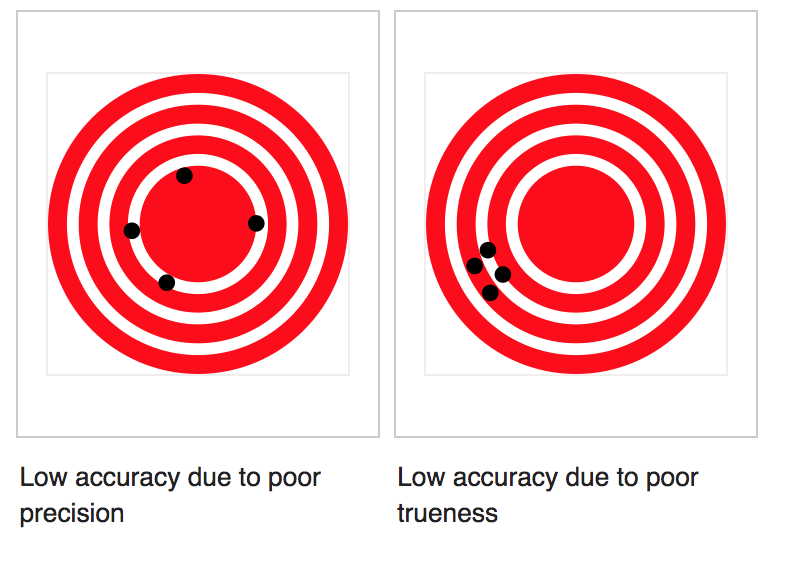}
\caption{Precision vs. Trueness. [Public domain], via Wikimedia Commons. (\url{http://bit.ly/2tzdPOI}).}
\end{figure}

\subsection{Interpretation Methods}
Concretely, we will compare the descriptive accuracy of the following feature explanation methods. We define a feature explanation method as the combination of a model and an interpretation algorithm that produces feature importance ranks. Two types of ranks can be produced. Local feature ranks represent which features are most impactful towards single prediction. Global feature ranks represent which feature impact the entire model the most.
Listed below are several popular interpretation algorithms.

\subsubsection{Linear Models}
Linear models are highly interpretable; feature importance ranks can be found by simply looking at the regression coefficients.
For example, the formula for OLS regression is as follows: \begin{equation}
Y = \beta_0 + \beta^\mathsf{T} X + \epsilon,
\end{equation} where $X$ is the feature matrix, $Y$ is the response, and $\epsilon$ represents irreducible error.
The regression coefficients $\beta_j$ can be interpreted as the change in the response $Y$ per unit change in a single feature $X_j$. If the feature matrix $X$ is scaled, the absolute value of the coefficient $\beta_j$ can be used to determine feature importance.
We will call this feature explanation method regression coefficient magnitude and abbreviate it as RCM.

\subsubsection{Additive Models}
Additive models such as Generalized Additive Models (GAM) and Microsoft Research's Explainable Boosted Machines (EBM) \cite{yinlouandcaruana2013accurate,1909.09223} provide their own feature explanations. Consider a standard GAM of the form  $\Sigma f_{i}(x_{i})$. For each feature $x_{i}$ the impact of the feature on the response is given by the term $f_{i}(x_{i})$. Calculating the impact of each instance in a dataset for a given feature, and averaging over all the instances, gives that feature's global importance.

\subsubsection{Tree-Based Mean Decrease Impurity}
A popular method for computing feature importance in tree models involves calculating the decrease in node impurity each time the tree splits on a feature \cite{2001.04295}. Averaging across all of the splits, in all of the trees,
of a model yields a score that ranks variable importance. This score is known as Mean Decrease Impurity (MDI). The larger a feature’s average decrease in node impurity per split, the more important the feature is towards the model. This method often used to evaluate feature importance in popular tree models such as Random Forests and Boosted Trees. 

\subsubsection{LIME}
Local Interpretable Model-Agnostic Explanations (LIME) is a feature explanation algorithm that fits localized explainer models to explain how individual predictions are made \cite{DBLP:journals/corr/RibeiroSG16}. The method is model agnostic; it does not depend on how the model is built, rather only on the predictions the model makes. The explainer models built during LIME are fit by solving
\begin{equation}
\arg\min_{g}  L(f,g,\pi_{x})+ \Omega(g),
\end{equation}
where $f$ is the model, $g$ is the explainer model and $\pi_{x}$ is the neighborhood around instance $x$. The loss function, $L(f,g,\pi_{x})$, measures of how closely the model and explainer model align in the neighborhood of $\pi_{x}$. The regularization function, $\Omega(g)$, penalizes complex explainer models to ensure that the explainer model fitted is easy to interpret. In practice, the explainer model $g$ is often selected from a family of linear models or decision trees due to their easy interpretability. It is important to note that LIME local explanations are not additive and can not be combined into a global feature importance score. Linden et al. found that when aggregated, LIME local explanations fail to reliably represent global model behavior \cite{1907.03039}.

\subsubsection{SHAP}
Similar to LIME, Shapley Additive Explanations (SHAP) is a feature explanation algorithm that fits a separate additive explanation model that is easier to interpret than the original \cite{DBLP:journals/corr/LundbergL17}. 
The explanation model used in SHAP is given by \begin{equation}
g(z') = \phi_{0} + \Sigma_{j=1}^{M}\phi_{i}z'_{i},
\end{equation} where $M$ is the number of imput features, $z'_{i}$ is a binary variable indicating the presence of feature $i$, and $z' \in \{0,1\}^{M}$. SHAP exploits the principle that there exists a unique solution for the feature attribution values $\phi_{i}$ that satisfies the properties of local accuracy, missingness, and consistency.

Local accuracy requires that the explanation model $g(z')$ match the original model $f(x)$ when the feature vector $x$ is mapped to the indicator vector $z'$. Missingness requires that features missing in the original input have no impact, and have $\phi_{i} = 0$. Consistency requires that if the marginalized contribution of a feature is larger in model $f$ compared to model $f'$, its attribution value $\phi_{i}(f,x)$ must be larger as well, i.e
\begin{equation} 
f(x) - f(x \setminus i)  \geq f'(x) - f'(x \setminus i) \implies \phi_{i}(f,x) \geq \phi_{i}(f',x).
\end{equation}

These unique feature attribution values (SHAP values) are calculated similarly to Shapley regression values and represent the average marginal contribution of a feature. Formally, we can calculate these values for a single instance of the dataset, x, by \begin{equation}
\phi_{i}(f,x) = \sum_{z' \subset x'} \frac{\lvert z'\rvert!(M -\lvert z'\rvert -1 )! }{M!} [f_{x}(z') - f_{x} (z' \setminus i)],
\end{equation}
where $M$ is the number of features, $x'$ is a binary vector that indicates the presence of features, and $z'$ ranges over all possible subsets of the entries of $x'$. This value gives the measure of local feature importance for feature $i$ and instance $x$. Unlike LIME, these local explanations explain global model behavior reliably when aggregated.

A complete list of the feature explanation methods we use is presented in the table 1.
\begin{table}[h]
\caption{Feature Explanation Methods}
\begin{tabular}{|l|l|l|}
\hline
\textbf{Model}      & \textbf{Interpretation Method} & \textbf{Type} \\ \hline
Logistic Regression &  Coefficient Magnitude      & Local and Global          \\ \hline
Random Forest       & Mean Decrease Impurity         & Global Only               \\ \hline
Random Forest       & SHAP                           & Local and Global          \\ \hline
Random Forest       & LIME                           & Local Only                \\ \hline
XGBoost             & Mean Decrease Impurity         & Global Only               \\ \hline
XGBoost             & SHAP                           & Local and Global          \\ \hline
XGBoost             & LIME                           & Local Only                \\ \hline
EBM                 & Model Itself                   & Local and Global          \\ \hline
\end{tabular}
\end{table}

\section{Methodology}
In this section, we present the methods we use to test how descriptive accuracy is affected by a model's predictive accuracy.
\subsection{Varying Predictive Accuracy}
To vary the predictive accuracy of a model, we first split our dataset of interest using a 70-30 train test split. The test split of the data is held constant while the train set is subsampled in decreasing proportions. This has the effect of reducing the amount of training data the model sees, which reduces the predictive accuracy of the model. Through this procedure, we can obtain accuracy vs. percent of training set used curves. Figure 2 shows such a curve when the above procedure is run on the Musk dataset \cite{Dua:2019}. The x-axis represents the percentage of the training set used and the y-axis represents the F1 score of the models.  We report F1 score as our accuracy metric to account for any potential class imbalance.

 \begin{figure}[h]
\includegraphics[width=8 cm]{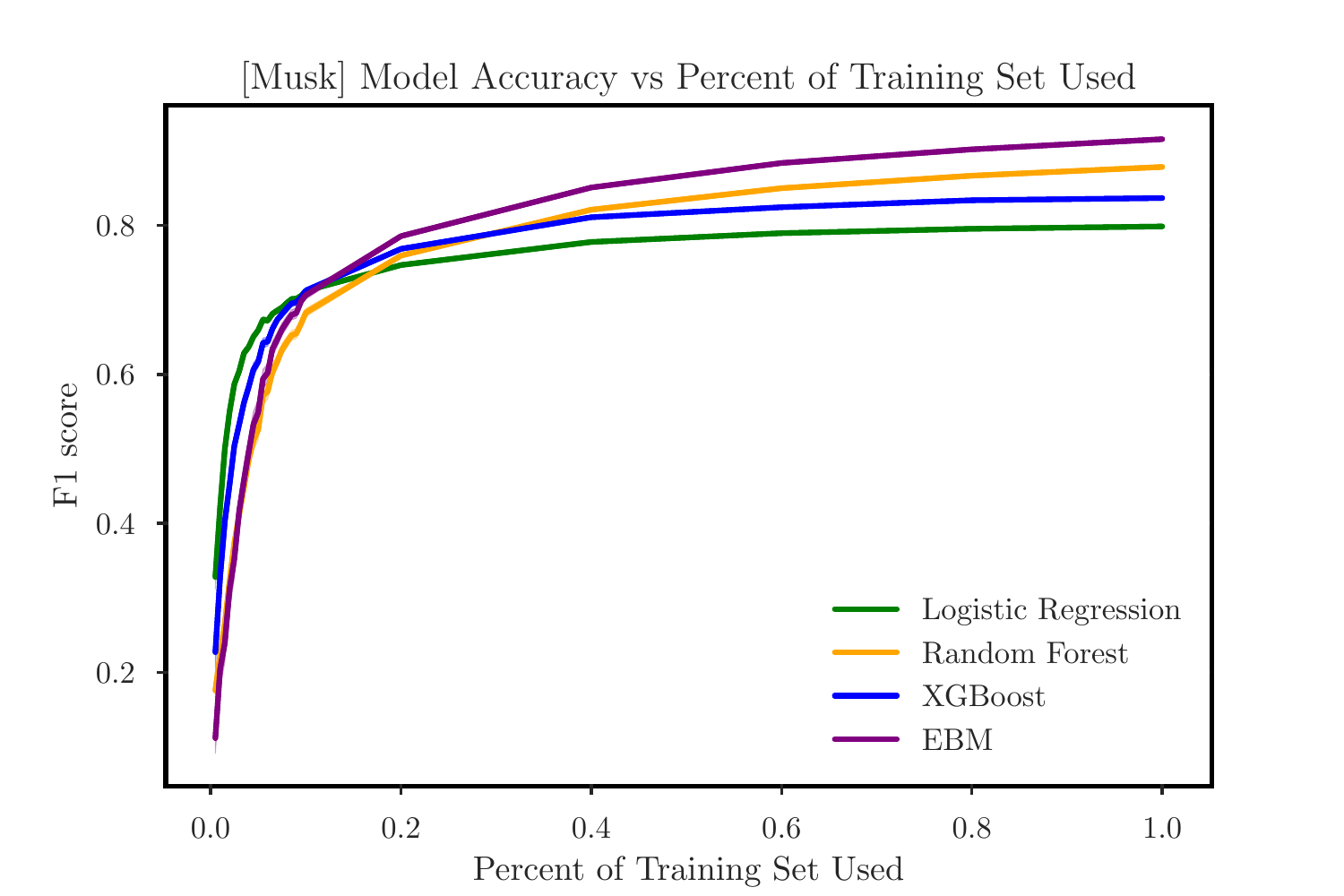}
\caption{Accuracy vs. Percent of Training Set Used}
\end{figure}
\subsection{Data Perturbations} 

To assess the stability of our interpretation methods, we perturb our data through bootstrapping. We bootstrap $N$ sample datasets at each proportion of the original data subsampled. We then fit models and explanation methods on each of the $N$ samples and assess the stability of the results. 

The algorithm for our procedure is presented as Algorithm 1.

\begin{algorithm}
\caption{Experimental Procedure}
\begin{algorithmic}
\STATE Given dataset $D \in \mathbb{R}^{M \times N}$ and set of feature explanations methods $F$
\FOR{$p$ in proportion range}
    \item bootstrap $D_{1},\ldots,D_{N}$ samples of size $p \times M$
    \FOR{$d \in D_{1},\ldots,D_{N}$}
        \FOR{$f \in F$}
            \item Evaluate feature explanation method $f$ on data sample $d$
            \item Record model accuracy and feature ranks
\ENDFOR{}
    \ENDFOR{}
        \ENDFOR{}
\end{algorithmic}
\end{algorithm}

\subsection{Bound on Interpretation Trueness}
We can formalize our bound on interpretation trueness as follows. Consider  dataset $D \in \mathbb{R}^{M \times N}$ and feature explanation method $f$ that outputs a vector of feature ranks $r \in  \mathbb{R}^{N}$. We are interested in the probability of obtaining the ground truth feature rank $r^{\star}$ when we apply $f$ once. In our procedure, we bootstrap our data to produce $D_{1},\ldots,D_{N}$ samples and repeatedly apply $f$ to obtain the set of feature rankings $r_{1},\dots,r_{N}$. For a given feature ranking $r_{i}$ we claim the the probability $r_{i}$ is the ground truth feature rank is boundedby the probability $r_{i}$ is equal to the mode. 

Let the mode of $r_{1},\dots,r_{N}$ be $r_{\rm{mode}}$. Consider the following cases:
\begin{itemize}
    \item Case 1: If the ground truth ranking coincides with the mode, $r^{\star} = r_{\rm{mode}}$, then  $P(r_{i} = r^{\star}) = P(r_{i} = r_{\rm{mode}})$.
    \item Case 2: If the ground truth ranking does not coincide the with the mode, then since the mode is defined as the most likely value for a set, $P(r_{i} = r^{\star}) \leq P(r_{i} = r_{\rm{mode}})$.
\end{itemize}

This upper bound is very small; it is exceedingly difficult to identify the full feature rankings. In practice it is not important to obtain the exact true feature rank from a model with hundreds of features. Rather, it is more important to correctly identify the top predictors in given model. Therefore, we will only consider the top 3 features in each ranking when computing the probability of obtaining the mode. We will present this probability as a metric for interpretation trueness and abbreviate it as pMode.

\subsection{Measuring Interpretation Stability}
To measure the stability of a set of feature rankings $r_{1},\dots,r_{N}$ we will compute the mean pairwise dissimilarity of the set.
To find the dissimilarity between to rankings $r_{i}$ and $r_{j}$ we will calculate the Kendall Tau distance between the two rank vectors \cite{kumar2010generalized}. This distance can be expressed as the minimum number of bubble swaps needed to convert on rank vector into another, divided by the total number of pairs in the vector. For example, the rank vectors ['A' 'B' 'C'] and ['B' 'A' 'C'] differ by a single bubble swap operation, and each rank vector contains $\binom{3}{2} = 3$ pairs, so the Kendall Tau distance between the ranks is equal to $\frac{1}{3}$. 

From a real world perspective, it is more important to identify the top predictors of a model, rather than accurately rank all the features. Therefore, we modify or Kendall Tau distance calculations to impose a $\frac{1}{k}$ penalty on any bubble swap involving the $k^{\rm{th}}$ rank. This ensures that our distance metric more heavily penalizes miss-alignments between rank vectors in the top few $(1^{\rm{st}}$, $2^{\rm{nd}}$, $3^{\rm{rd}})$ ranks. In addition, we truncate the rank vectors after the top 10 ranks to reduce computation time

With our new weighted Kendall Tau metric, we can calculate the distance between each pair of vectors in $r_{1},\ldots,r_{N}$. The mean of these $\binom{N}{2}$ pairwise distances represents the stability of our rank measurements. We invert the value of our final metric so that ranks that are entirely consistent have a score of 1 and entirely opposite ranks have a score of 0. We abbreviate this metric as WKT10.

\section{Experiment}
We test our experimental procedure on 25 datasets commonly used for binary classification benchmarking. On each dataset, we run our experiment for $N = 200$ trials.

\subsection{Datasets}
All the datasets we considered are based on real-world data and were found in repositories such as UCI MLR \cite{Dua:2019} and KEEL \cite{journals/mvl/Alcala-FdezFLDG11}. They span a diverse range of domains, from healthcare to manufacturing, and sizes, in both number of instances and features. Table 2 presents a sample of the datasets used.

\begin{table}[h]
\caption{Datasets Used}
\begin{tabular}{|l|l|l|}
\hline
\textbf{Dataset}               & \textbf{Instances} & \textbf{Features} \\ \hline
Appendicitis                    & 106                & 7                 \\ \hline
Bupa Liver Disorder             & 345                & 7                 \\ \hline
Car Evaluation                  & 1728               & 6                 \\ \hline
Cardiotocography                & 2130               & 40                \\ \hline
Chess (King-Rook vs. King-Pawn) & 3196               & 47                \\ \hline
Congressional Voting Records    & 435                & 16                \\ \hline
Contraceptive Choice            & 1473               & 9                 \\ \hline
Dishonest Internet Users        & 322                & 5                 \\ \hline
Glass Identification            & 214                & 10                \\ \hline
Ionosphere                      & 351                & 34                \\ \hline
Musk                            & 6598               & 170               \\ \hline
Phishing Websites Data          & 2456               & 30                \\ \hline
Phoneme                         & 5404               & 5                 \\ \hline
Somerville Happiness Survey     & 143                & 7                 \\ \hline
\end{tabular}
\end{table}

\subsection{Implementation}
We implement our experiment in Python and use open source packages (sklearn, lime, shap, xgboost, glassbox) to fit each feature explanation method. The experiments are conducted on a Google Cloud Compute VM with 96 cores and 86 GB memory.

\section{Results}
When we run our experiments on individual datasets, we can plot our interpretation stability/trueness metrics against model accuracy. Figure 3 gives an example of such a plot. The graph shows the results of our experiment run on the Japanese Credit Screening dataset. The x-axis of the plot shows the F1 score of our classification models and the y-axis shows our interpretation stability metric for global feature rankings. The error bands are perturbation intervals \cite{yu2019three} that give a sense of how the metric varies when the data is perturbed. Through this plot, we can compare the impact of predictive accuracy on descriptive accuracy for feature explanation methods fit on a single dataset. To compare results across all datasets, we discretize accuracy and compute the average interpretation stability/trueness metric of each range. The procedure is detailed below.

\begin{figure}[h]
\includegraphics[width=8 cm]{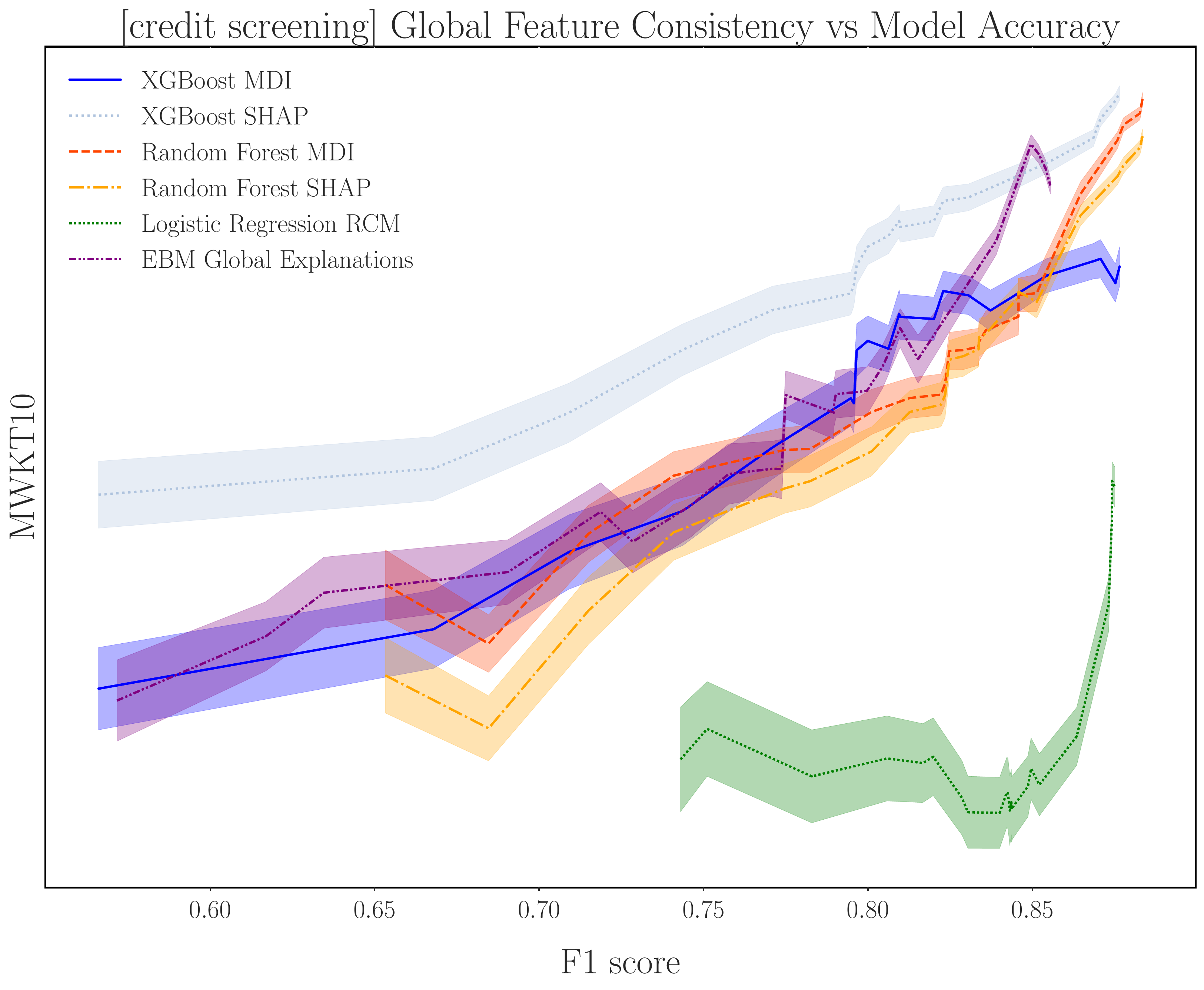}
\caption{Credit Screening: Global Interpretation Stability vs. Accuracy}
\end{figure}

 \begin{figure*}[h]
\includegraphics[width=8 cm, height = 8 cm]{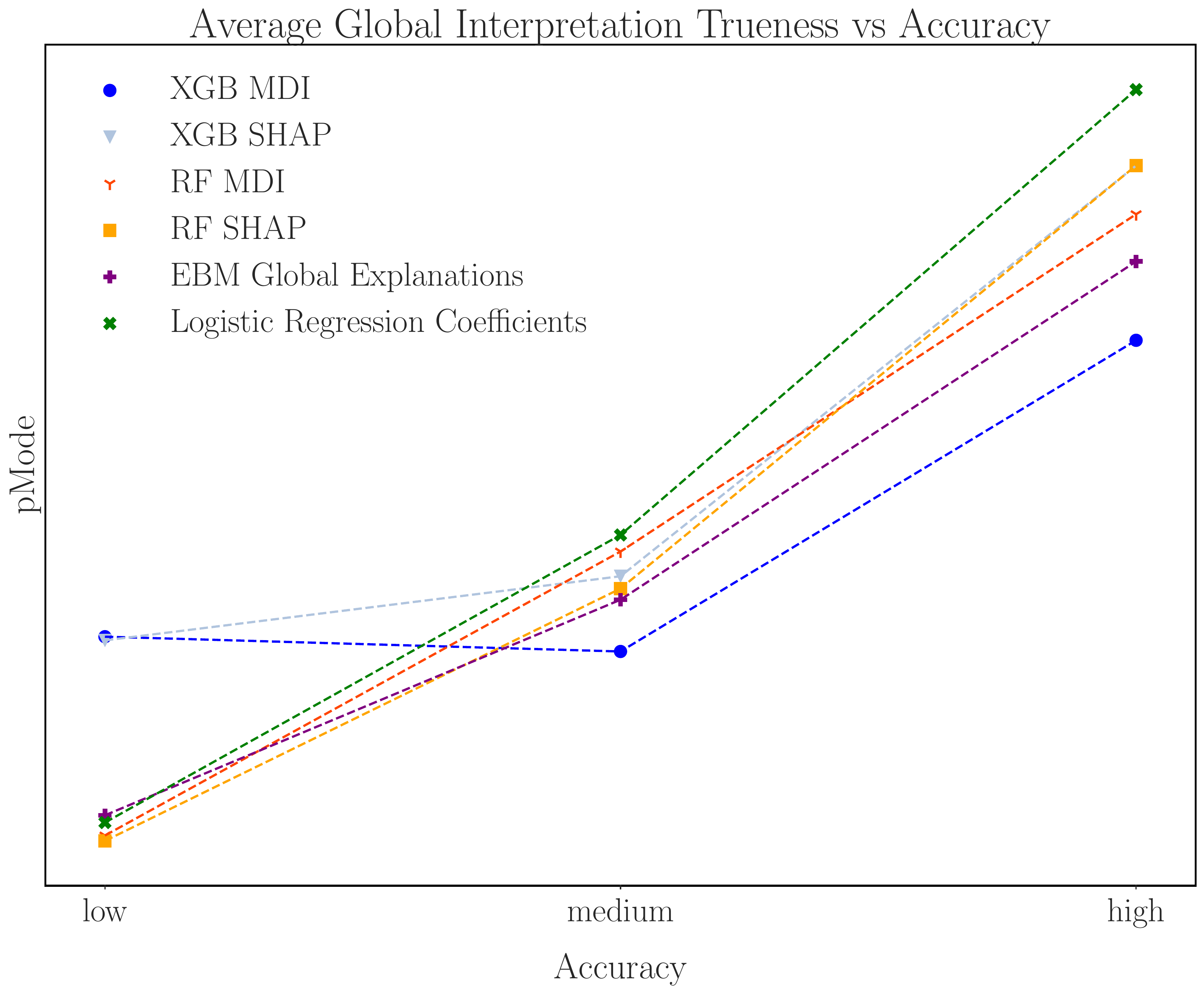}
\includegraphics[width=8 cm, height = 8 cm]{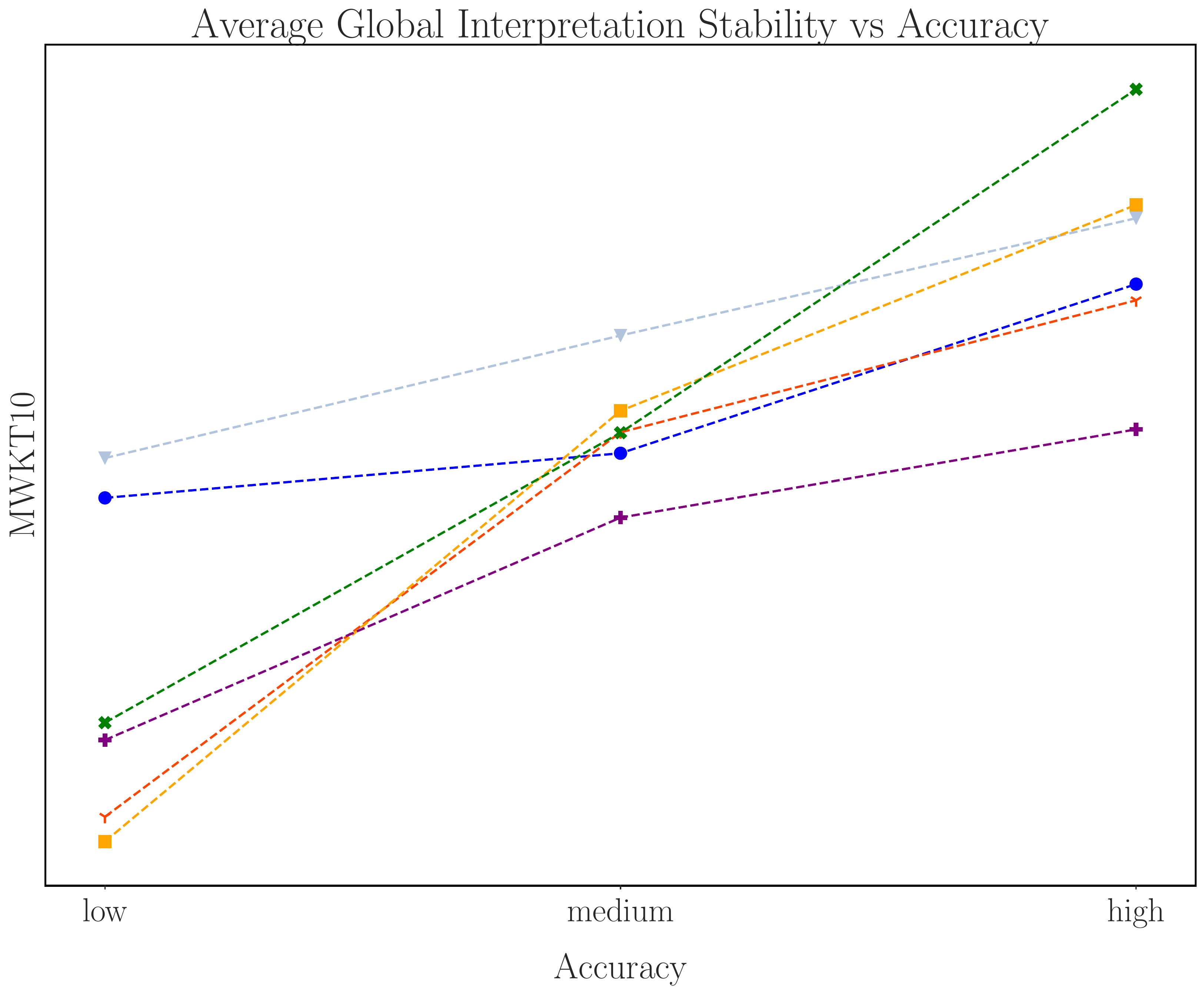}
\includegraphics[width=8 cm, height = 8 cm]{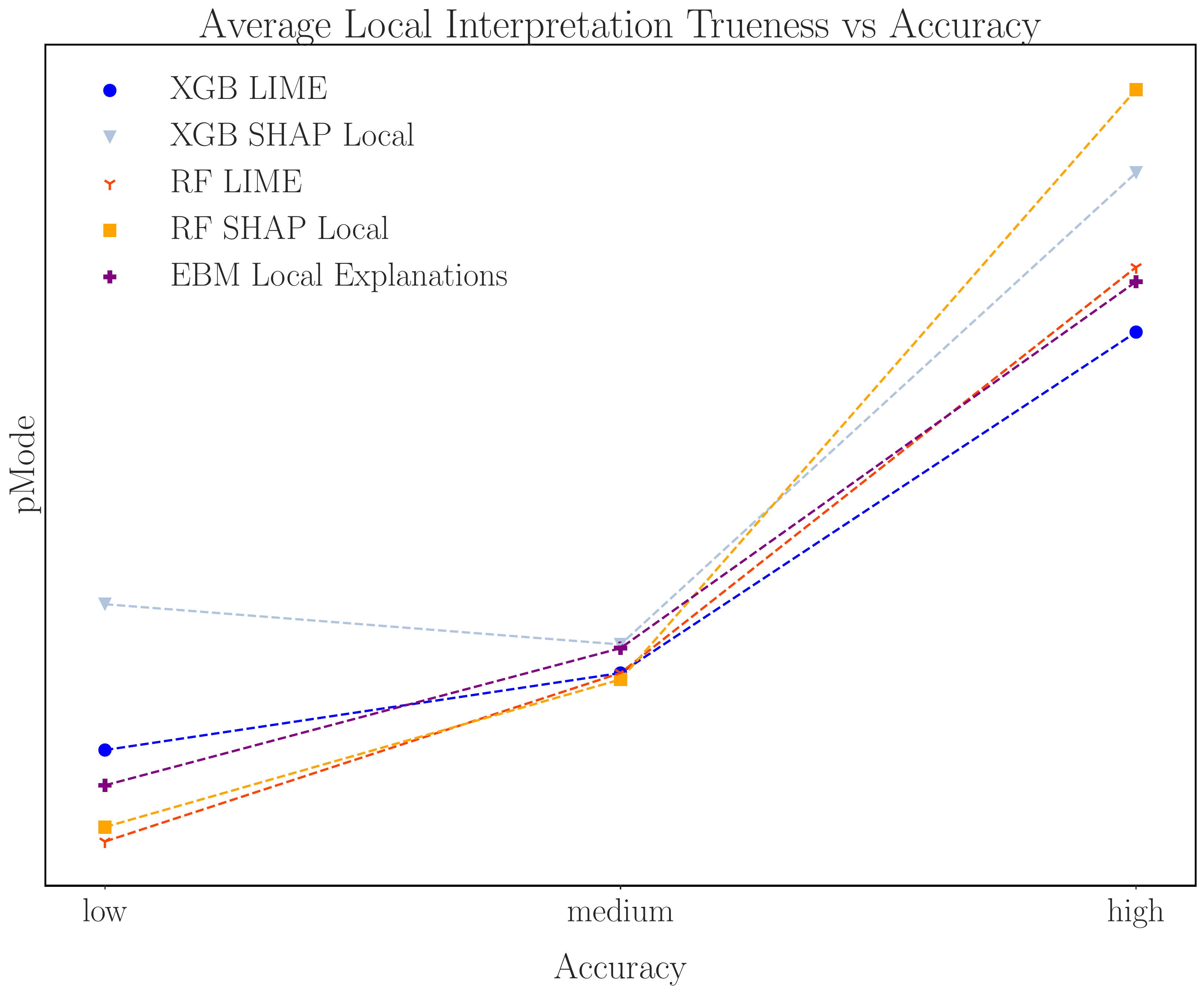}
\includegraphics[width=8 cm, height = 8 cm]{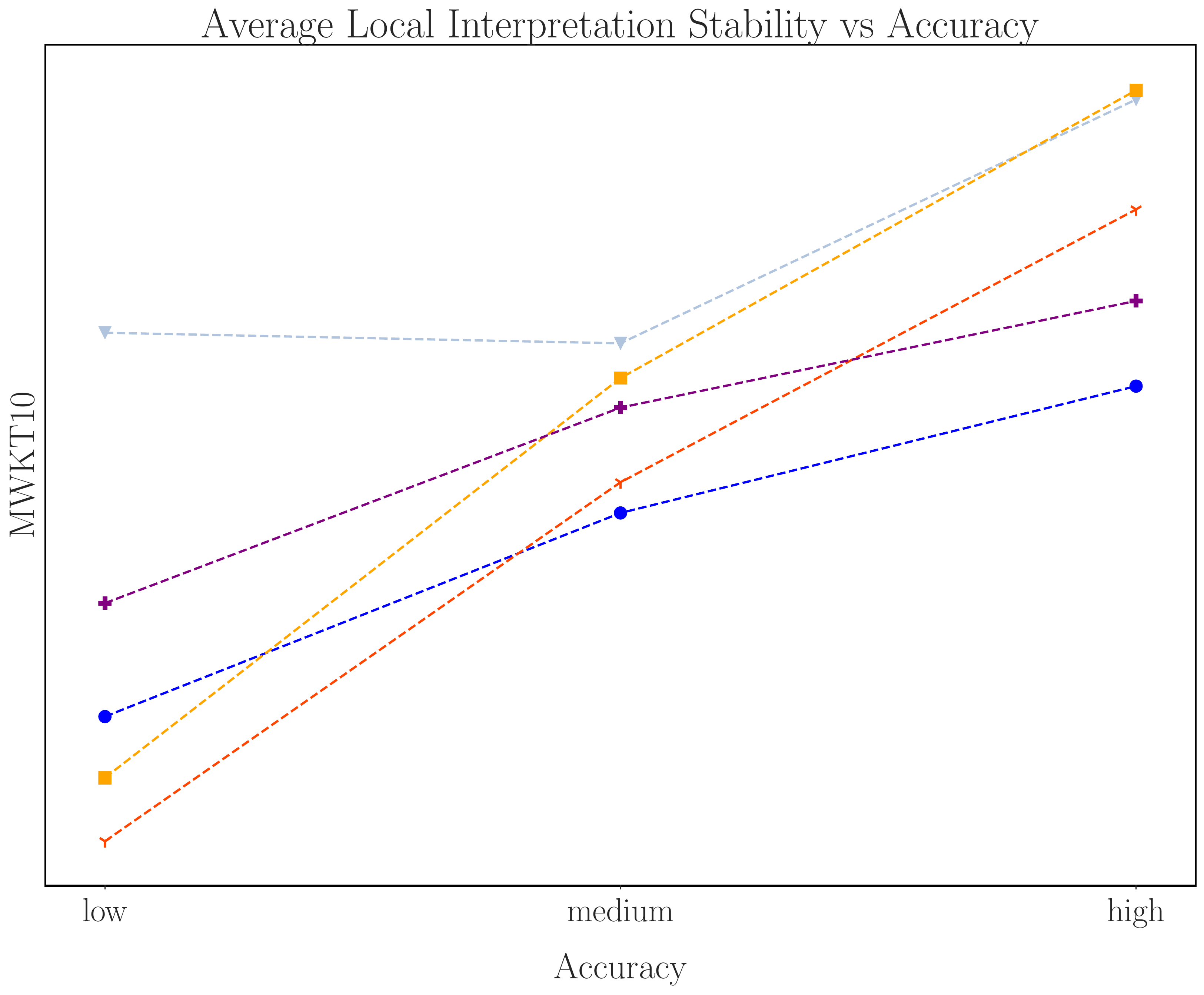}
\caption{Average Metrics vs. Accuracy}
\end{figure*}

\subsection{Accuracy Discretization}
We discretize accuracy into 3 distinct buckets. The predictive accuracy of a model is defined as low if the F1 score of the model is between $[0.5, 0.65]$, medium if between $[0.65, 0.8]$, and high if between $[0.8, 1.0]$. For each accuracy bucket, we record the interpretation stability and trueness for each feature explanation method fit on each dataset. We then compute the mean metric of each accuracy bucket. The results are shown in figure 4. 

The top two plots in figure 4 show that for global feature rankings, XGBoost with SHAP performs the best at low levels of accuracy. This feature explanation method outperforms all of the others in terms of average interpretation stability and trueness.
At medium levels of accuracy, XGBoost with SHAP still performs the best in terms of interpretation stabilty. However, Logistic Regression RCM performs slightly better in terms of interpretation trueness. At high levels of accuracy, Logistic Regression RCM performs the best for both metrics.

For local feature rankings, the bottom two plots in figure 4 show that XGBoost with SHAP again performs the best at low levels of accuracy. At medium levels of accuracy, EBM Local Explanations outperform XGBoost with SHAP in terms of interpretation trueness.
At high levels of accuracy, Random Forest with SHAP outperforms XGBoost with SHAP for both metrics.

 \begin{figure*}[h]
\includegraphics[width=16 cm,height = 10 cm]{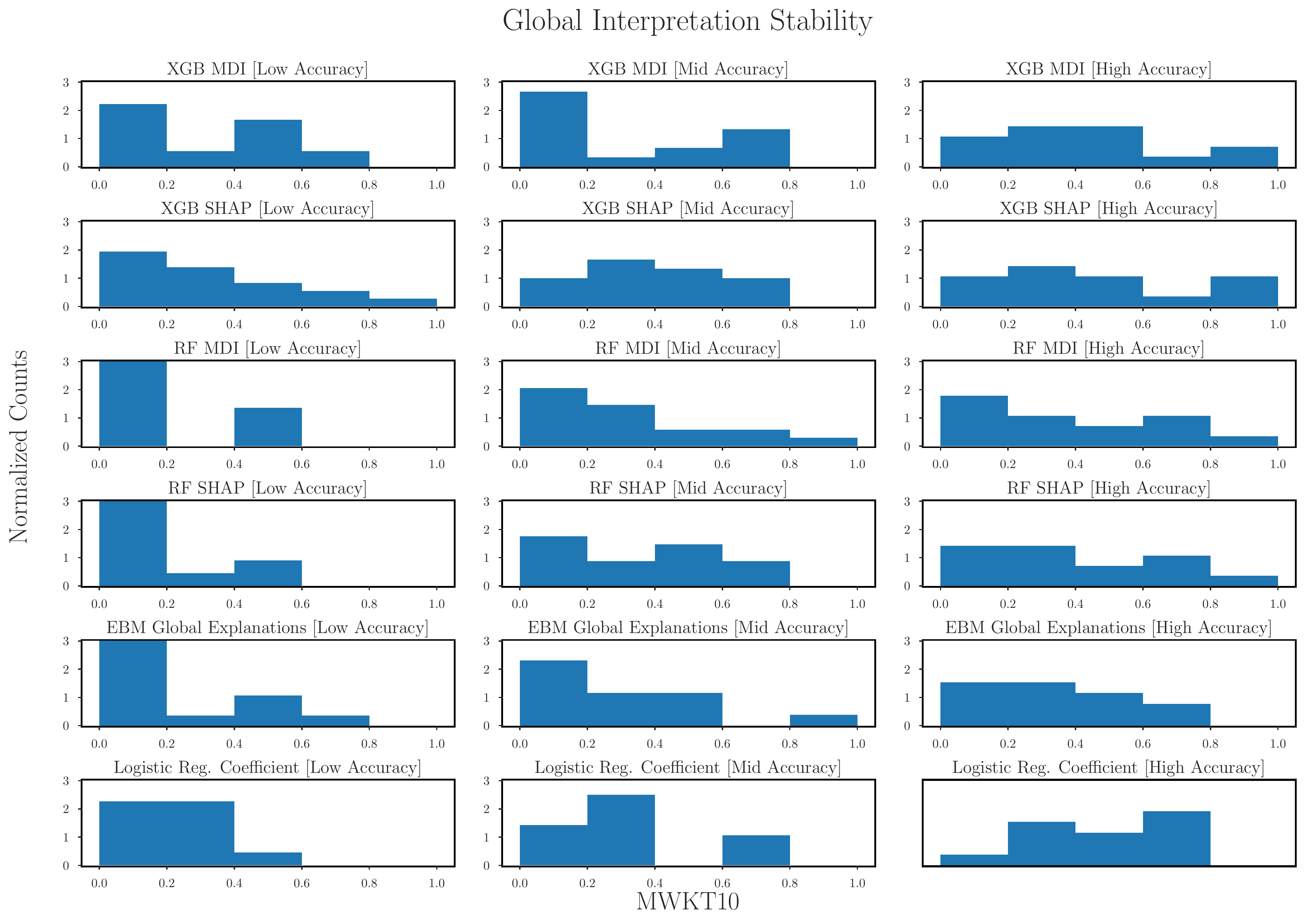}
\caption{Distribution of Global Interpretation Stability}
\end{figure*}

In figure 5 we present histograms of the normalzed distribution of global feature importance stability over all the datasets used. In the histogram matrix, each row represents a different feature explanation method and each column represents an accuracy bucket.
One interesting result can be seen in the second row of figure 5. This row contains the normalized distribution of our interpretation stability metric for XGB SHAP global feature explanations. For the low and medium accuracy buckets, the distributions are fairly unimodal. Other feature explanation methods mostly have bimodal interpretation stability distributions in these two buckets. This provides further evidence that XGBoost SHAP is more stable than other global feature explanation methods when model accuracy is low.

\subsection{Analysis}

In figure 4, we observe that as model accuracy increases, interpretation quality increases for both global and local feature explanations. When we plot interpretation stability/trueness against accuracy at the dataset level, such as in figure 3, we see this behavior as well.

For both global and local feature explanations, XGBoost with SHAP is the best performing method in the low accuracy bucket. As model accuracy increases, Logistic Regression RCM performs best for global feature explanations and Random Forest with SHAP performs best for local feature explanations. For global explanations, Logistic Regression RCM is a much simpler feature explanation method than XGBoost with SHAP. XGBoost is substantially more complex than logistic regression, and applying SHAP is much more complicated than ranking the magnitudes of regression coefficients. For local explanations, the boosting algorithm used in XGBoost is more complex and less stable than the bagging algorithm used in Random Forests.  

This result indicates that if multiple models achieve high levels of predictive accuracy, using a simpler model/interpretation algorithm may yield higher quality interpretations. However, when the predictive accuracy of all models is low, XGBoost with SHAP may be the best choice to obtain trustworthy interpretations.

We also see in figure 4 that for global feature explanations, SHAP outperforms MDI interpretations when applied to both XGBoost and Random Forests. This is consistent with Lunberg's findings that SHAP produces more consistent global feature explanations when compared to tradition methods such as MDI and split-count \cite{lundberg2018consistent}.

In addition, for local feature explanations, SHAP always outperforms LIME when used with either model. This may be due to the fact that SHAP is a more stable algorithm than LIME. SHAP relies on averaging the marginal contributions of a feature, while LIME involves perturbing an instance and measuring the response \cite{1907.03039}. This additional stability may result in more trustworthy interpretations.

\section{Recomendations}
We make the following recommendations to data scientists and analysts on which feature explanation methods perform the best under certain circumstances.
\begin{itemize}
    \item If the dataset is easy to model and all models attain high predictive accuracy, use simpler feature explanation methods for high quality interpretations.
    \item If the dataset is difficult to model and predictive accuracy is low, XGBoost with SHAP has the best change of producing trustworthy interpretations.
    \item When using popular tree algorithms such as XGBoost and Random Forests, use SHAP for global and local feature explanations rather than MDI interpretations and LIME.
\end{itemize}

Additionally, we recommend that data scientists use our procedure of subsampling and bootstrapping to test how their feature explanations vary when model accuracy is perturbed. Reporting perturbation intervals, such as the error bands in figure 3, may help data scientists convince others that their interpretations are trustworthy.

\section{Conclusion}

In summary, this paper proposes two new metrics (pMode an WKT10) that quantify an interpretation's descriptive accuracy, presents an experiment that uses subsampling to test how interpretation quality varies with model accuracy, and presents a list of recommendations of which feature explanations perform best in different situations. We conclude that for datasets that can be modeled accurately by a variety of methods, simpler feature explanations work the best. For datasets that are difficult to model, XGBoost with SHAP provides the most trustworthy interpretations. By better understanding how predictive accuracy impacts interpretation quality, we hope to influence how data scientists decide what makes a trustworthy interpretation.

\subsection{Future Work}
Future work can be done testing our experimental procedure on tasks other than binary classification, such as regression and text/image analysis. This may help generalize our results to other fields in data science and machine learning. Additionally, analysis can be done on how the size of a dataset impacts the relationship between predictive accuracy and descriptive accuracy.

\bibliographystyle{ACM-Reference-Format}
\bibliography{bibfile}

\end{document}